\documentclass[letterpaper]{article} 
\usepackage{aaai2026}  
\usepackage{times}  
\usepackage{helvet}  
\usepackage{courier}  
\usepackage[hyphens]{url}  
\usepackage{graphicx} 
\usepackage{natbib}  
\usepackage{caption} 
\usepackage{algorithm}
\usepackage{algorithmic}

\usepackage{amsmath,amsfonts}
\usepackage{amssymb}
\usepackage{amsthm}
\usepackage{makecell}
\usepackage{booktabs}
\usepackage{multicol}
\usepackage{multirow}
\usepackage{diagbox}

  \usepackage[caption=false,font=footnotesize]{subfig}

\usepackage{newfloat}
\usepackage{subcaption} 
\usepackage{color}
\usepackage{listings}
\DeclareCaptionStyle{ruled}{labelfont=normalfont,labelsep=colon,strut=off} 
\floatstyle{ruled}
\newfloat{listing}{tb}{lst}{}
\floatname{listing}{Listing}
\title{TRACE: A Generalizable Drift Detector for Streaming Data-Driven Optimization}
\author {
    Yuan-Ting Zhong\textsuperscript{\rm 1},
    Ting Huang\textsuperscript{\rm 2},
    Xiaolin Xiao\textsuperscript{\rm 3},
    Yue-Jiao Gong\textsuperscript{\rm 1}\footnote{Corresponding author}
}
\affiliations {
    \textsuperscript{\rm 1}South China University of Technology\\
    \textsuperscript{\rm 2}Xidian University\\ 
     \textsuperscript{\rm 3}South China Normal University\\
    \{ytalienzhong, gnauhgnith, shellyxiaolin, gongyuejiao\}@gmail.com
}

\usepackage{bibentry}

\begin{document}

\maketitle

\begin{abstract}
Many optimization tasks involve streaming data with unknown concept drifts, posing a significant challenge as Streaming Data-Driven Optimization (SDDO). Existing methods, while leveraging surrogate model approximation and historical knowledge transfer, are often under restrictive assumptions such as fixed drift intervals and fully environmental observability, limiting their adaptability to diverse dynamic environments. We propose \textbf{TRACE}, a \underline{TRA}nsferable \underline{C}oncept-drift \underline{E}stimator that effectively detects distributional changes in streaming data with varying time scales. TRACE leverages a principled tokenization strategy to extract statistical features from data streams and models drift patterns using attention-based sequence learning, enabling accurate detection on unseen datasets and highlighting the transferability of learned drift patterns. Further, we showcase TRACE's plug-and-play nature by integrating it into a streaming optimizer, facilitating adaptive optimization under unknown drifts. Comprehensive experimental results on diverse benchmarks demonstrate the superior generalization, robustness, and effectiveness of our approach in SDDO scenarios.
\end{abstract}

\vspace{-0.5cm}
\begin{links}
    \link{Code}{https://github.com/YTALIEN/TRACE}
\end{links}
\vspace{-0.5cm}

\section{Introduction}
Many real-world optimization applications are driven by massive volumes of continuously arriving data.
For instance, traffic optimization in smart cities relies on real-time data streams from sensors and monitoring systems~\cite{styler2015real, kang2019dynamic,ji2022stden}. 
In streaming environments, the underlying data distribution may change unpredictably over time due to external factors such as traffic accidents or weather fluctuations, which is a phenomenon known as concept drift~\cite{gower2025identifying}.
These drift occurrences, often unknown and associated with limited data at each time step, give rise to Streaming Data-Driven Optimization (SDDO) problems~\cite{zhong2024sddobench}, where optimization strategies must adapt dynamically to maintain performance.

Recent efforts to address SDDO problems have led to the development of Streaming Data-Driven Evolutionary Algorithms (SDDEAs)~\cite{richter2020model,yang2023data,zhang2024solving}. These algorithms offer a promising approach by integrating evolutionary optimization techniques with data-driven modeling.  
Typically, SDDEAs build surrogate models from streaming data and transfer knowledge from past environments to accelerate optimization of the current environment~\cite{luo2018surrogate}.
Although these approaches have demonstrated encouraging results, their performance often hinges on strong, and often unrealistic assumptions. 
For example, some methods assume fixed and known drift intervals, which allows them to explicitly adapt the optimization strategy at the beginning of each interval~\cite{li2023datadriven,zhang2024solving}. 
Others assume immediate access to complete data from each environment before making optimization decisions~\cite{yang2023data,liu2025data}.
In more complex scenarios with unpredictable drifts and continuous data streams, the absence of reliable drift detection can lead SDDEAs to overfit outdated distributions or overlook recurring patterns, ultimately degrading optimization performance.

\begin{figure}[t]
    \centering
    \includegraphics[width=0.9\linewidth]{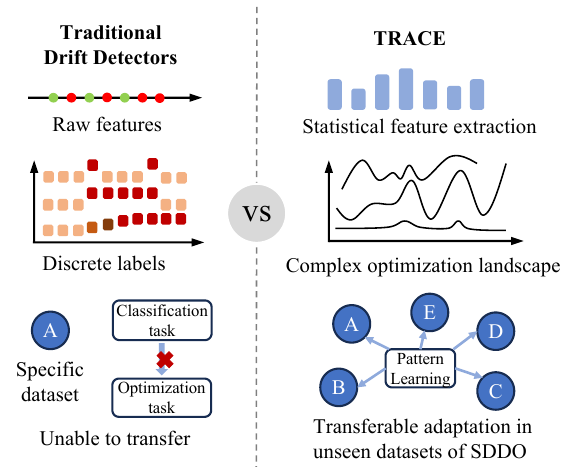}
    \caption{TRACE offers a learnable and generalizable approach to drift detection in the SDDO landscape.}
    \label{fig:TRACE-comparison}
\end{figure}

Despite increasing interest in SDDO, effective and generalizable drift detection methods tailored for streaming optimization remain scarce. While drift detection has been explored in stream data mining~\cite{alsaedi2023radar,wan2024online}, existing methods are primarily designed for classification tasks, which struggle with the unique characteristics of SDDO. 
Specifically, they often assume discrete labels or bounded outputs, which are incompatible with the unbounded, real-valued domains common in SDDO. 
Furthermore, these methods typically focus on detecting abrupt changes in prediction accuracy, overlooking the subtle performance degradation that often signifies drift in optimization landscapes. 
To bridge this gap, we propose \textbf{TRACE}, a \underline{TRA}nsferable \underline{C}oncept-drift \underline{E}stimator that flexibly solves streaming data with varying time scales and distribution shifts, offering a tailored solution for SDDO environments. 
As illustrated in Figure~\ref{fig:TRACE-comparison}, 
TRACE demonstrates a strong generalization ability, enabling it to detect drift in previously unseen datasets by learning and leveraging transferable drift patterns.
Moreover, TRACE can be seamlessly integrated in many SDDEAs as a general-purpose detector, enabling adaptive optimization under unknown concept drift.
Our main contributions are summarized as follows:
    \textbf{1) Principled Stream Tokenization for Drift Modeling:} We introduce a tokenization strategy to transform streaming data into a sequence of statistical representations, capturing temporal distributional characteristics that are indicative of concept drift. 
    The statistical sequences serve as informative inputs for drift detection models, enabling pattern understanding and facilitating scalable, label-efficient training of supervised detection models. 
    \textbf{2) Unified Framework for Learning Transferable Drift Patterns:}  Building upon the statistical representation mentioned above, we design a generalizable detection framework based on attention-driven sequence modeling. Our attention mechanism captures local and global temporal dependencies, which, together with training on diverse changing landscapes, prevents overfitting to specific data streams.
    Our framework allows the model to learn transferable drift patterns, ensuring robust and accurate detection across diverse tasks and previously unseen datasets.
    \textbf{3) Plug-and-Play Detection Module for Streaming Optimizers:} 
    TRACE can be seamlessly integrated in many SDDEAs as a general-purpose detector, without requiring modifications to the optimization logic.
    To demonstrate this, we design TRACE-EA, a SDDO algorithm that embeds TRACE as a drift-awareness module.
    This flexible design demonstrates a pathway for enabling SDDEAs to effectively operate under streaming environments with unknown concept drift, proactively responding to distributional changes.
    \textbf{4) Comprehensive Validation across Tasks and Domains:} 
    We conduct extensive experiments to demonstrate the effectiveness, generalization ability, and practical utility of our approach in diverse streaming scenarios.
    Our evaluation covers unseen datasets, cross-task optimization, ablation analysis, and real-world applications, showcasing its robustness and broad applicability.

\section{Related Work}

\textbf{SDDEAs.} 
SDDEAs address SDDO challenges by reusing knowledge from past environments to accelerate the current environmental optimization. 
\textit{Solution-level reuse} is common: SAEF-1GP~\cite{luo2018surrogate} augments training data with the optimal solution from most recent environment.
SAEA-TL~\cite{wu2024surrogate} reuses historical data whose surrogate yields the lowest RMSE in the current environment. 
DETO~\cite{li2023datadriven} clusters past surrogate parameters and injects solutions from the closest cluster.
\textit{Model-level reuse} methods include DSEMFS~\cite{yang2023data}, which aggregates surrogates from a maintained surrogates' pool, and MLDE~\cite{zhang2024solving} employs meta-learning to initialize new models with parameters from previous ones. 
While offering promising directions for SDDO, these SDDEAs commonly assume fixed drift intervals and full data availability per environment. This leads to treating each data batch as a distinct environment, which fails to capture the unpredictable nature of real-world data streams.
DASE~\cite{zhong2025data} takes a step forward by incorporating statistical drift detection (HCDD) into streaming optimization, but its fixed-threshold tests limit adaptability. 
This fundamental mismatch between method assumptions and the characteristics of real-world streaming data limits the effectiveness of current SDDEAs.
As a result, their practical utility is restricted in applications that demand continuous, online adaptation.

\textbf{Drift Detection in Stream Data Mining.}
Stream data mining~\cite{alsaedi2023radar,wan2024online} has seen the development of numerous drift detection methods, broadly categorized as statistical or learning-based.
Among \textit{statistical methods}, DDM~\cite{gama2004learning} monitors increases in prediction error means, while HDDM\_A/W~\cite{frias-blanco2015online} improve adaptivity via Hoeffding's and McDiarmid's inequalities.
Two-window methods compare statistics between recent and historical data: ADWIN~\cite{bifet2007learning} detects significant mean differences using Hoeffding's bound; KSWIN~\cite{raab2020reactive} applies the Kolmogorov–Smirnov test, and PUDD~\cite{lu2025early} introduces a PU-index using the Chi-Square test for theoretically grounded detection.
\textit{Learning-based methods} have gained increasing interest recently:
RADAR~\cite{alsaedi2023radar} uses a recurrent variational embedding to learn latent dynamics and detects drift in the representation space.
MCD-DD~\cite{wan2024online} leverages contrastive learning to estimate maximum concept discrepancy between sample pairs, enabling robust detection in dynamic environments.
However, these methods face key limitations in SDDO with regression-type objectives and surrogate models: 1) Designed for classification, they assume discrete labels or bounded outputs, hence unsuitable for SDDO's unbounded, real-valued domains.
2) In optimization, drift may manifest not as prediction error spikes, but as performance degradation stemming from evolving objective landscapes or shifting surrogate model behavior. 
Those methods struggle to distinguish such phenomena.
3) Threshold-based methods relying on hand-crafted rules often lack generalizability across different scenarios. 
The challenges highlight the pressing need for more flexible, generalizable, and adaptive drift detection mechanisms tailored to SDDO.

\section{Methodology}
We begin by introducing a tokenization strategy to transform streaming data into sequences suitable for drift modeling. 
We then detail the architecture and training procedure of TRACE, our proposed transferable concept-drift estimator.
Finally, we present TRACE-EA, an instantiation of SDDEA integrates TRACE as a plug-and-play module for adaptive SDDO problems with unknown concept drift.

\subsection{Stream Tokenization for Drift Modeling}
To effectively model concept drift in streaming data, we first need to transform the raw data stream into a format suitable for our drift detection model. 
This involves a process we refer to as stream tokenization, in which the continuous stream is converted into a labeled sequence of discrete tokens that capture the underlying dynamics. 
We explore a statistical approach, detailed below.

Given a stream $\{(\mathbf{x}_1,y_1),\cdots,$ $(\mathbf{x}_i,y_i),\cdots\}$, where $\mathbf{x}_i$ represents the sample $i$ and $y_i$ is the corresponding objective value, we train a surrogate model for each environment to predict the objective value given the input. 
Then, we compute the prediction error of the surrogate for sample $i$ as:
\begin{equation}
\label{eq:er}
    e_i=\begin{cases}
             \left|(y_i-\hat{y}_i)/y_i\right|, \; & y_i \neq 0 \\
             |y_i-\hat{y}_i|, \; & \text{otherwise}
         \end{cases}
\end{equation}
where $\hat{y}_i$ is the surrogate-predicted objective value.

In static environments, prediction errors fluctuate around a stable mean.  
However, when drift occurs, changes in the data distribution lead to a degradation in the surrogate model's predictive performance, often manifesting as noticeable change in prediction error. These patterns are what we aim to capture.
Our goal is to create labeled sequences that can be used to train a supervised drift detection model. We achieve this through a two-step process: 1) extracting statistical features from the prediction errors within sliding windows to create tokens, and 2) combining these tokens into labeled sequences that incorporate environmental context and temporal coherence.

First, we apply a sliding window of length 
$n$ to capture the current statistical properties of prediction errors. 
At time $t$, window $t$ contains $n$ most recent data samples: $\left\{\left(\mathbf{x}_{t-n+1}, y_{t-n+1}\right), \ldots,\left(\mathbf{x}_t, y_t\right)\right\}$, by which we compute the prediction errors $\left\{e_{t-n+1}, \ldots, e_t\right\}$ using Eq.(\ref{eq:er}). 
From these, we extract a statistical feature vector: $fv_t=(\mu_t,\sigma_t,{\min}_t,{\max}_t,Q1_t,Q2_t,Q3_t)$,
where $\mu_t$ and $\sigma_t$ denote the mean and standard deviation, and $Q1_t$, $Q2_t$, and $Q3_t$ are the first, second (median), and third quartiles of the error distribution within window $t$.
Applying a sliding window over the stream yields a sequence of feature vectors.

To create training samples, we select $T$ consecutive feature vectors: $\{fv_1,fv_2,\cdots,fv_T\}$.
To incorporate environmental context, we prepend a special context token $fv_0$, computed over all data observed in the current environment up to the start of the $T$ windowed vectors, summarizing the environment's state.
Each training sample is then formed as a sequence of $(T+1)$ tokens: $\left<fv_0,fv_1,fv_2,\cdots,fv_T\right>$, with an associated drift labeled $dl$. 
If no drift occurs within the subsequence, $dl = 0$; otherwise, $dl = l$ indicates drift occurs at the $l$-th step ($1 \leq l \leq T$).
Thus, each sample is a labeled token sequence $X \in \mathbb{R}^{(T+1) \times d_f}$, where $d_f$ is the dimensionality of each feature vector.

\subsection{TRACE for Learning Transferable Drift Patterns}

\begin{figure*}[h]
    \centering
    \includegraphics[width=\linewidth]{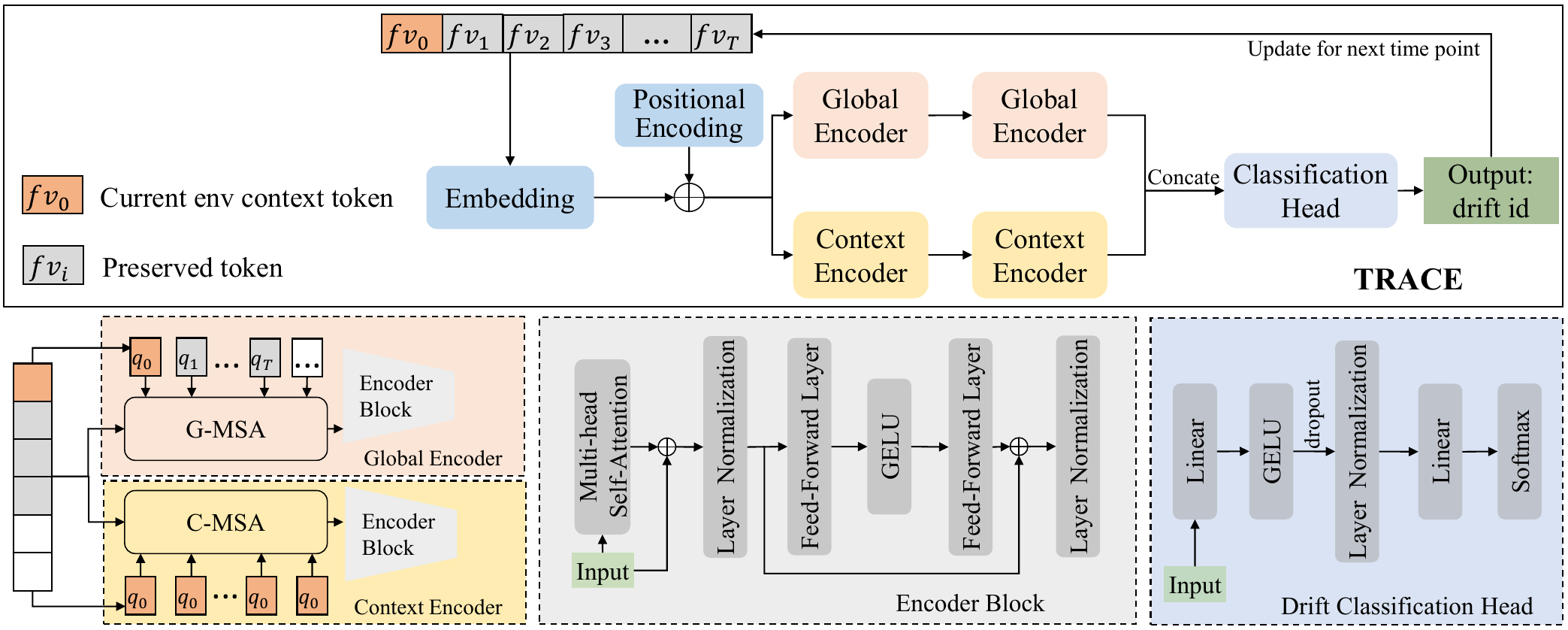}
    \caption{TRACE Architecture. The upper part illustrates the overall framework, while the lower part details its key components:  the Sequence Embedding Module;  the Dual-Attention Encoder; and  the Drift Classification Head.}
    \label{fig:TRACE-arch}
\end{figure*}

Leveraging labeled sequences derived from prediction errors, we unlock rich representations that implicitly capture evolving drift patterns in data streams - a feat often unattainable with raw data alone.  
Conventional statistical detectors struggle with complex temporal and structural changes, especially under sudden, incremental, or mixed drifts.  
To overcome these challenges, we propose TRACE, a learning-based drift estimator that extracts discriminative, transferable representations directly from sequences of window-level features.  
As illustrated in Figure~\ref{fig:TRACE-arch}, TRACE comprises three key components:
1) an embedding module that maps input features into a high-dimensional latent space,
2) an encoder that models temporal dependencies and structural patterns across the sequence, and
3) a classification head that predicts the drift label from encoded representations.
The following sections detail each component of TRACE.

\subsubsection{Sequence Embedding.}
Streaming data results in input sequences of variable lengths. To handle this, sequences are padded to a fixed maximum length using special \texttt{<PAD>} tokens that the model learns to ignore. Embedding and positional encoding then transform these statistical feature sequences into temporally aware representations, enabling the model to capture semantic relationships and sequential dependencies. 
Further details are provided in Appendix A\footnote{``Appendix.pdf'' in our codebase.}.

\subsubsection{The TRACE Encoder.}

TRACE's encoder employs two complementary self-attention mechanisms to capture diverse temporal dependencies critical for drift detection:

\textit{1) Global Multi-Head Self-Attention (G-MSA)}: 
G-MSA models global interactions among all tokens, capturing long-range temporal and structural patterns indicative of concept drift. This allows the model to capture diverse interaction patterns across the sequence.
The G-MSA follows the standard Transformer encoder~\cite{vaswani2017attention}, enabling each token to attend to every other token, capturing long-range dependencies indicative of drift, which often signal the emergence of concept drift.
With the integration of positional encoding, G-MSA is sensitive to relative order of tokens, thus modeling both temporal and structural patterns.
Formally, given embedded inputs with positional encoding $\mathbf{H}^{\text{pos}}$, G-MSA first calculates queries $\mathbf{Q}$, keys $\mathbf{K}$, and values $\mathbf{V}$ using learned projection matrices $\mathbf{W}_Q$, $\mathbf{W}_K$, and $\mathbf{W}_V$. 
For each attention head $i$, the attention output is $\text{h}^G_i=\text{Softmax}\Big(\mathbf{Q}_i\mathbf{K}_i^{\top}/\sqrt{d_k}\Big)\mathbf{V}_i$. Finally,  the outputs from $m$ heads are concatenated and projected via $\mathbf{W}_{\rm{G}}$: $\operatorname{G-MSA}(\mathbf{H}^{\text{pos}})=\operatorname{Concat}(\text{h}^G_1,\cdots,\text{h}^G_m)\mathbf{W}_{\rm{G}}$.

\textit{2) Context-Guided MSA (C-MSA)}: 
A crucial insight for drift detection is measuring how recent data deviates from the current environment.
C-MSA directly addresses this by modeling the relationship between each token and a dedicated context token that encodes the current environment. 
Recall that the first token in the input sequence represents the most recent data distribution and serves as a reference for detecting distributional changes.
To highlight deviations relative to this context, C-MSA uses the context token as the sole query: $\mathbf{q} = \mathbf{h}_0 \mathbf{W}_q$, while treating all other tokens as keys and values, following the same formulation as G-MSA.
This design explicitly models the relationship between the current environment context and recent tokens, effectively capturing incremental drift patterns.

TRACE combines G-MSA and C-MSA to model both global temporal dependencies and localized context-aware deviations.
The concatenated outputs form a joint representation that encodes structural patterns and environment-specific changes.
This dual-attention encoder enables TRACE to generalize across diverse optimization tasks, with minimal assumptions about drift characteristics.

\subsubsection{Drift Classification Head.}
The output of the dual-attention encoder is fed into a pointer-like classification head. This head predicts the drift index within the input sequence or indicates the absence of drift.
This formulation enables both explicit localization of drift events and confirmation of distributional stability.

Specifically, the classification head consists of two linear layers with GELU activation, followed by dropout and layer normalization to enhance regularization and training stability.
It outputs a probability distribution over $T+1$ classes: $T$ classes correspond to the time windows in the sequence, and the extra class indicates the no-drift case.
Formally, given the concatenated embedding $\mathbf{z}$ from G-MSA and C-MSA, the drift classification head is:
\begin{equation}
\mathbf{y}=\text{Softmax}(\mathbf{W_2}\cdot \text{LayerNorm}(\phi(\mathbf{W}_1\cdot\mathbf{z}+\mathbf{b}_1))+\mathbf{b}_2)
\end{equation}
where $\mathbf{y} \in \mathbb{R}^{T+1}$ represents the probabilities over all windows ${1, \dots, T}$ and the no-drift class (label 0).

\subsubsection{Training for Transferability.}

To provide a controlled environment for learning,
TRACE is trained using synthetic streams from SDDObench~\cite{zhong2024sddobench}, a dedicated benchmark for SDDO, building on prior work in benchmarking dynamic optimization problems~\cite{li2008benchmark,yazdani2022benchmarking}.  
SDDObench simulates diverse types of concept drifts, allowing TRACE to learn from varied drift and non-drift scenarios in a controlled yet rich setting.
Training mimics a realistic streaming process: starting with a surrogate model built for the initial environment.
As the stream progresses, labeled sequences are generated via a sliding window to represent recent distributional characteristics. 
When new data signals a shift, the surrogate model is updated for the current optimization landscape. 
To improve generalization, training samples are randomly truncated after the true drift index, exposing TRACE to diverse temporal patterns.
The model is optimized using cross-entropy loss over the predicted drift index, promoting both detection and accurate localization. 
This setup enables TRACE to generalize well to new datasets and unseen drift conditions.

\subsection{Integration into Streaming Optimizers}

\subsubsection{The TRACE-EA Framework.}
TRACE is designed as a plug-and-play drift detection module compatible with many SDDEAs.
To demonstrate its effectiveness, we develop TRACE-EA by integrating TRACE into DASE~\cite{zhong2025data} as the drift detector, replacing DASE's original hand-crafted drift detection mechanism, namely the HCDD.
TRACE-EA addresses three challenges in SDDO problems: accurately detecting distribution changes, rapidly adapting optimization without restarting, and enabling transferable drift detection across unseen tasks.
The following part details the detection-adaptation loop that governs TRACE-EA's continuous operation. A more detailed description of TRACE-EA is provided in the Appendix B.

\subsubsection{The Detection-Adaptation Loop.}
TRACE-EA operates continuously via a detection-adaptation loop integrating streaming data processing, drift detection, and adaptive optimization.
At each time $t$, the algorithm receives a small data batch $\mathbb{D}_t$, updates the sliding window, and constructs token sequences as input to TRACE.
If no drift is detected, optimization proceeds normally.
Upon drift detection, a new environment is instantiated, and the archive is queried to identify similar past environments based on feature similarity. 
A knowledge transfer module then reuses relevant surrogate models and population knowledge to warm-start the optimization in the new environment. Here, a core component is an archive-based knowledge transfer mechanism. Instead of discarding prior optimization information when environment changes, TRACE-EA maintains an archive of past environments and learned optimization knowledge. When a drift is detected, it identifies the most relevant prior environment using similarity metrics and selectively transfers surrogate models and population states.
After adaptation, the archive is updated with new knowledge. 
This continuous detection-adaptation loop allows TRACE-EA to maintain high performance in dynamic environments by rapidly responding to changes and avoiding redundant re-optimization.
By reusing experience to guide adaptation, TRACE-EA achieves efficient and robust performance under streaming data environments.

\section{Experiments}
Our evaluation addresses the following research questions:
\textbf{RQ1}: How well does TRACE generalize to unseen datasets for accurate drift detection?
\textbf{RQ2}: Can TRACE-EA effectively solve optimization problems unseen during training?
\textbf{RQ3}: What is the contribution of each core component of TRACE to the drift detection performance?
\textbf{RQ4}: How does TRACE-EA perform on real-world stream clustering tasks?

\subsection{Experimental Setup}

\subsubsection{Training setup.}  SDDObench~\cite{zhong2024sddobench} is used to generate training data. For each instance in SDDObench, we create 60 environments, each containing a randomly chosen number of samples from $\{600, 750, 900\}$.
The sliding window size $n$ is set to 30, and sequences have a maximum length of 20. 
A Radial Basis Function Network (RBFN) serves as the surrogate model for errors computation~\cite{zhong2025data}.
Training is performed with batch size 32, a fixed learning rate of $5 \times 10^{-4}$, over 50 epochs.
All experiments run on a machine with an AMD EPYC 9745 CPU @ 3.45GHz, and NVIDIA RTX 4080 Super GPU, using Python 3.10.12 and PyTorch 2.0.1.

\subsubsection{Competitors.}
We consider two groups of competitors.
First, for drift detection only, TRACE is compared with established drift detectors from different fields, including  DDM~\shortcite{gama2004learning}, ADWIN~\shortcite{bifet2007learning}, 
HDDM\_A/W~\shortcite{frias-blanco2015online}, FHDDM~\shortcite{pesaranghader2016fast},  KSWIN~\shortcite{raab2020reactive}, RADAR~\shortcite{alsaedi2023radar}, MCD\_DD~\shortcite{wan2024online}, and HCDD~\shortcite{zhong2025data}. 
Second, for the higher-level SDDO tasks, our TRACE-EA is compared against several state-of-the-art SDDEAs, including
SAEF-1GP~\shortcite{luo2018surrogate}, 
BDDEA-LDG~\shortcite{li2020boosting}, 
TT-DDEA~\shortcite{huang2021offline}, 
DSEMFS~\shortcite{yang2023data}, DETO~\shortcite{li2023datadriven},
MLDE~\shortcite{zhang2024solving}, and
DASE~\shortcite{zhong2025data}. 
All baselines are obtained from publicly available source code or faithfully re-implemented based on their original papers.

\subsubsection{Generalization Tests.}
To evaluate the generalization capabilities of TRACE, we conduct experiments using both In-Distribution (ID) and Out-Of-Distribution (OOD) datasets. For ID, we utilize \textit{SDDObench}~\cite{zhong2024sddobench} that employs a distinct set of problem instances generated with configurations explicitly disjoint from those used during training.
For OOD, we conduct experiments on \textit{DBG}~\cite{li2008benchmark} and \textit{GMPB}~\cite{yazdani2022benchmarking} to evaluate TRACE's performance on unseen distributions. In addition, we also evaluate TRACE on four real-world data stream clustering datasets to provide a realistic assessment.

\subsubsection{Performance Metrics.}
Standard metrics are used: 
\textit{Precision} and \textit{F1} for drift detection accuracy, while the \textit{Dynamic Tracking Error} ($E_{DT}$) for the average deviation from the optimum over time.
All results are averaged over 11 independent runs.
Statistical significance is determined using the Kruskal–Wallis test~\cite{kruskal1952use} followed by Dunnett's post-hoc analysis~\cite{dunnett1955multiple}. 
More experimental details are in Appendix C.

\begin{figure}[t]
    \centering
    \vspace*{0.1cm}
    \let\theparentsubfig\relax 
    \mbox{\includegraphics[width=0.9\linewidth]{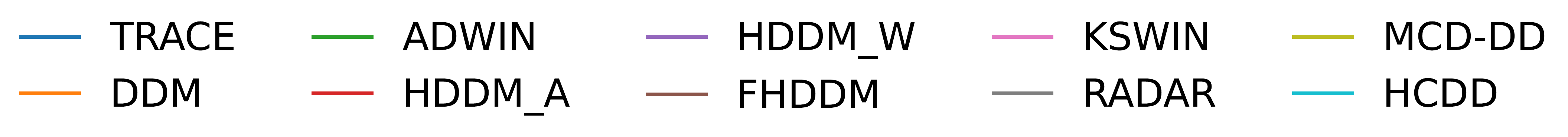}} 
    \par\vspace{-3mm}
    \subfloat[SDDObench\_F4]{\includegraphics[width=0.4\linewidth]{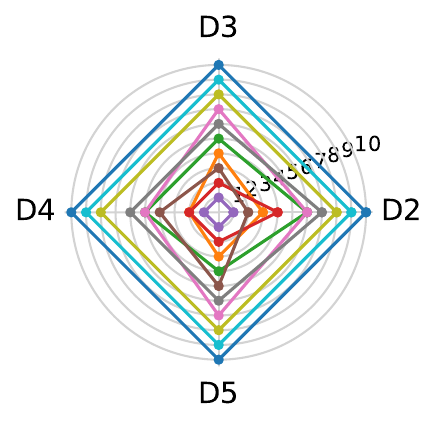}
    }
    \subfloat[DBG\_F1]{
    \includegraphics[width=0.4\linewidth]{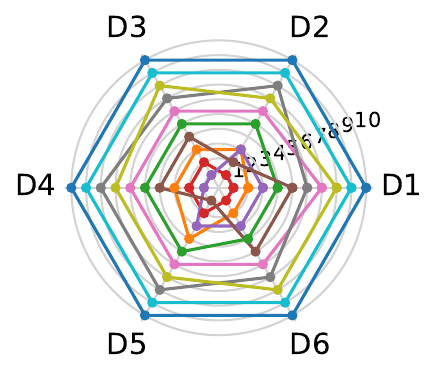}
    
    }
    \par\vspace{-0.4cm}
    \subfloat[DBG\_F4]{
    \includegraphics[width=0.4\linewidth]{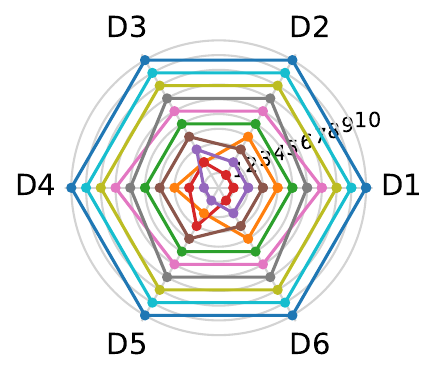}
    }
    \subfloat[GMPB]{
    \includegraphics[width=0.4\linewidth]{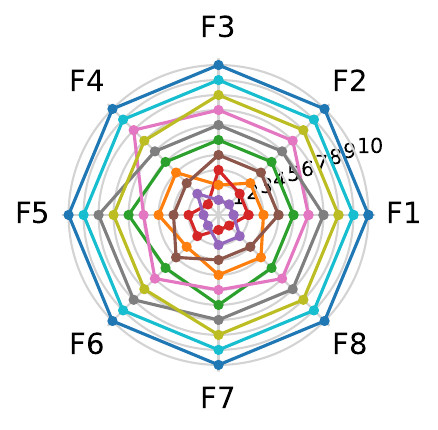}
    }
    \caption{Detector performance comparison. 
    See Appendix D-1 for additional results.}
    \label{fig:detector-comparison}
    \vspace{-2mm}
\end{figure}

\begin{table*}[t]
    \centering
    
    \scriptsize 
    \renewcommand{\arraystretch}{0.8}
    \setlength{\tabcolsep}{2pt}
    \resizebox{\textwidth}{!}{%
    \begin{tabular}{l|c|c|c|c|c|c|c|c}
        \hline
        Instance & TRACE-EA & SAEF-1GP & BDDEA-LDG & TT-DDEA & DSEMFS & DETO & MLDE & DASE \\ 
        \hline
        (a) F4D1 & \textbf{9.9e-02$\pm$3.7e-02} & 4.9e+01$\pm$2.2e-01 & 4.4e+00$\pm$2.1e+00 & 6.3e+00$\pm$4.5e-01 & 6.4e+01$\pm$1.1e+00 & 1.6e+02$\pm$2.2e+00 & 1.1e+02$\pm$1.8e+01 & 2.7e-01$\pm$1.7e-01 \\ 
        (a) F4D2 & \textbf{2.3e+00$\pm$2.0e-01} & 5.8e+01$\pm$1.0e+00 & 2.6e+01$\pm$3.2e+00 & 6.5e+01$\pm$1.4e+01 & 4.4e+01$\pm$1.1e+00 & 1.4e+02$\pm$9.0e-01 & 9.5e+01$\pm$6.4e+00 & 2.4e+01$\pm$1.6e-01 \\ 
        (a) F4D4 & \textbf{1.4e+01$\pm$1.5e-01} & 6.8e+01$\pm$9.6e-01 & 4.9e+01$\pm$4.2e-01 & 1.2e+02$\pm$6.3e+01 & 6.0e+01$\pm$2.4e+00 & 1.0e+02$\pm$5.8e+00 & 6.4e+01$\pm$6.4e+00 & 1.6e+01$\pm$2.9e-01 \\ 
        (b) F1D1 & \textbf{5.2e+01$\pm$1.7e+00} & 6.3e+01$\pm$2.0e+00 & 6.2e+01$\pm$9.6e-02 & 5.9e+01$\pm$2.2e-02 & 5.3e+01$\pm$6.3e-02 & 6.0e+01$\pm$2.9e+00 & 6.2e+01$\pm$1.1e+00 & 5.9e+01$\pm$2.0e+00 \\ 
        (b) F1D2 & \textbf{3.9e+01$\pm$6.4e-01} & 6.3e+01$\pm$2.5e-02 & 6.2e+01$\pm$9.6e-02 & 5.9e+01$\pm$2.2e-02 & 6.3e+01$\pm$8.4e-02 & 6.2e+01$\pm$3.7e-02 & 5.9e+01$\pm$2.3e+00 & 5.9e+01$\pm$9.1e-01 \\ 
        (b) F1D6 & \textbf{3.7e+01$\pm$1.4e-01} & 5.4e+01$\pm$3.9e-02 & 5.1e+01$\pm$1.5e-02 & 5.3e+01$\pm$1.4e+00 & 5.3e+01$\pm$1.1e-01 & 5.7e+01$\pm$5.7e-01 & 4.7e+01$\pm$9.7e-01 & 5.0e+01$\pm$6.1e-01 \\ 
        (c) F1 & \textbf{8.2e+02$\pm$1.1e+01} & 1.5e+03$\pm$1.4e+01 & 1.0e+03$\pm$2.5e+00 & 1.2e+03$\pm$4.5e+01 & 1.1e+03$\pm$1.0e+01 & 1.1e+03$\pm$1.1e+01 & 1.2e+03$\pm$4.5e+01 & 1.3e+03$\pm$4.1e+01 \\ 
        (a) F5 & \textbf{8.6e+02$\pm$7.9e+00} & 9.9e+03$\pm$4.8e+01 & 1.7e+03$\pm$2.7e+01 & 1.0e+03$\pm$3.3e+02 & 1.7e+03$\pm$4.1e+01 & 1.1e+03$\pm$4.2e+01 & 1.1e+03$\pm$2.7e+02 & 1.1e+03$\pm$4.1e+02 \\ 
        (a) F8 & \textbf{8.6e+02$\pm$7.4e+00} & 1.5e+03$\pm$6.3e+01 & 2.5e+03$\pm$1.0e+02 & 1.1e+03$\pm$2.6e+02 & 2.2e+03$\pm$7.7e+01 & 1.0e+03$\pm$1.4e+01 & 1.1e+03$\pm$1.6e+02 & 9.7e+02$\pm$2.6e+01 \\ 
        \hline
    \end{tabular}%
    }
    \vspace{-2mm}
    \caption{Comparison of $E_{DT}$ values among SDDEAs across benchmarks. See Appendix D-2 for all results.}
    \label{tab:sddea-compare}
    \vspace{-6mm} 
    \hrule height 0pt 
    \begin{minipage}{\linewidth}
        \centering
        \vspace{2mm} 
        \subfloat{
            \includegraphics[width=0.48\linewidth]{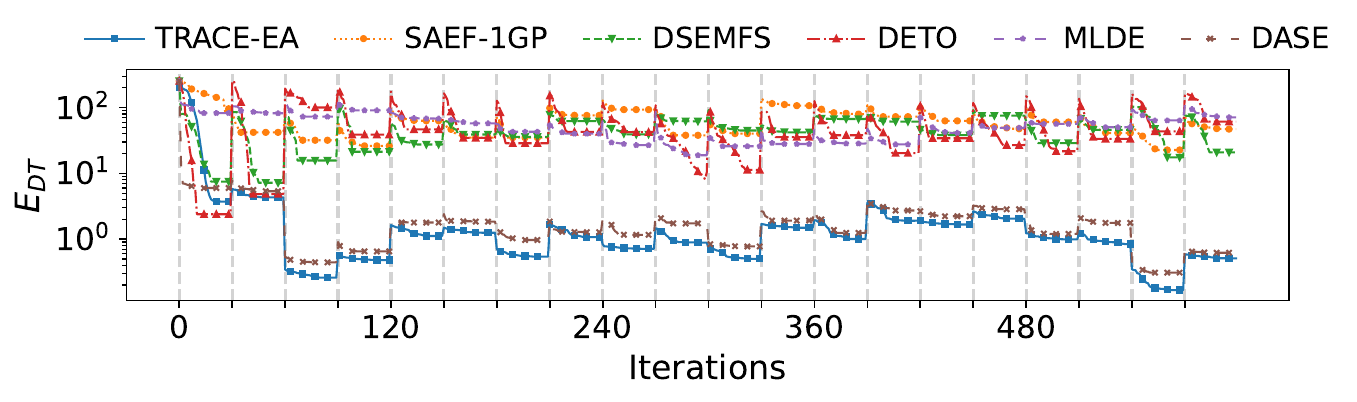}
        }
        \hfill 
        \subfloat{
            \includegraphics[width=0.48\linewidth]{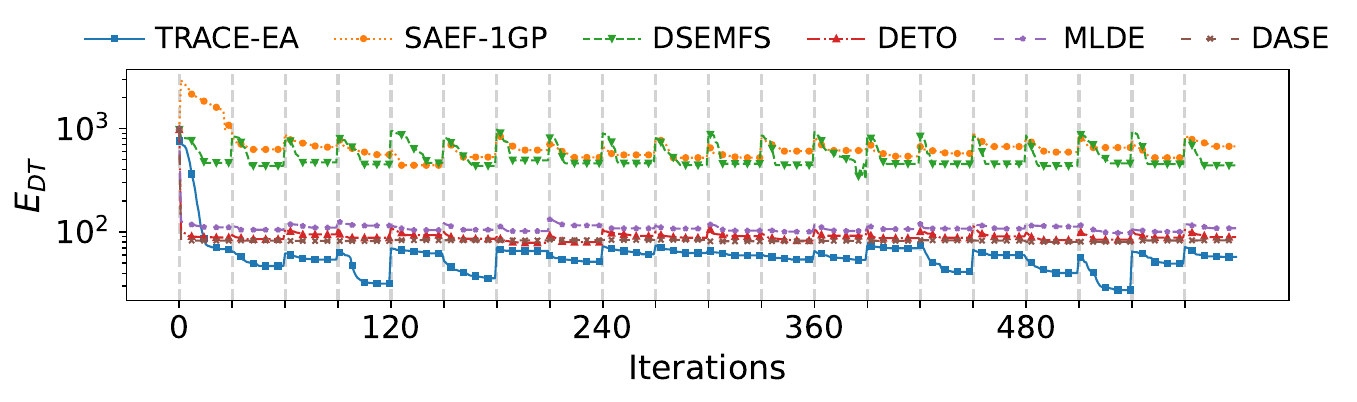}
        }
        \vspace{-3mm}
        \captionof{figure}{Convergence trajectories over the first 20 environments. Left: SDDObench\_F4D4; Right: DBG\_F1D2.}
        \label{fig:conver}
    \end{minipage}

    \vspace{-3mm} 
\end{table*}

\subsection{Detector Performance Comparison (RQ1)}

Figure~\ref{fig:detector-comparison} presents the precision on a subset of benchmarks; complete results across all benchmarks are available in the Appendix D-1.
TRACE consistently outperforms all baselines, achieving the highest precision on most tasks. 
It remains robust under both incremental and sudden drift scenarios (SDDObench\_F1: D2, D4), where traditional methods often suffer from high false positive.
Moreover, TRACE demonstrates strong performance on OOD datasets (GMPB and DBG), indicating effective transferability.
These results highlight TRACE's ability to capture generalizable representations of concept drift, enabling accurate detection across varying drift types.
Overall, TRACE not only surpasses statistical-based methods in accuracy but also generalizes reliably to diverse, unseen tasks.

\subsection{SDDO Performance Comparison (RQ2)}

Table~\ref{tab:sddea-compare} reports the average and standard deviation of $E_{DT}$ values obtained in SDDO on a subset of benchmarks; complete results across all datasets are available in Appendix D-2. 
TRACE-EA consistently outperforms all baselines across diverse benchmarks and drift types, highlighting its strong generalization beyond the training distribution. 
Its advantage stems from accurate drift detection via TRACE and effective reuse of historical knowledge, including archived populations and surrogate models, to initialize optimization in new environments. 
Concretely, as shown in Figure~\ref{fig:conver}, in the complex optimization landscape of SDDObench\_F4 with incremental drift (D4), TRACE precisely detects the onset of drift, enabling the algorithm to adapt quickly and prevent performance degradation.
This capability mitigates data scarcity and enhances efficiency under streaming environments.
TRACE also demonstrates a \textbf{fast detection response and low computational overhead}. Further experimental results are presented in Appendix D-4.

\subsection{Ablation and In-Depth Analysis (RQ3)}
\subsubsection{Ablation Study.} We examine the contribution of each component to TRACE's performance.
Specifically, we introduce four variants: 1) \textit{w/o PE}: Remove positional encodings; 2) \textit{w/o G-MSA}: Removes G-MSA; 3) \textit{w/o C-MSA}: Removes C-MSA; 4) \textit{vanilla class}: Replaces the pointer-like classifier with a standard fully connected layer. 
Table \ref{tab:ablation} summarizes the results. 
The removal of G-MSA causes a sharp performance drop, underscoring its role in capturing long-range patterns. 
Eliminating C-MSA also degrades accuracy, indicating its importance in environment-aware token interpretation. 
Replacing the pointer-like classification head with a vanilla classifier reduces performance, highlighting the advantage of explicit drift localization. 

\begin{table}[t]
    \centering
    \renewcommand\arraystretch{1}
    \resizebox{\linewidth}{!}{
    \begin{tabular}{c|cc|cc|cc|cc}
\hline
Instance & \multicolumn{2}{c|}{D1} & \multicolumn{2}{c|}{D2} &\multicolumn{2}{c|}{D3} & \multicolumn{2}{c}{D4}  \\ \hline
Metric & Prec & F1 & Prec & F1 & Prec & F1 & Prec & F1 \\ \hline
\textbf{TRACE }& \textbf{0.77} & \textbf{ 0.73} & \textbf{0.75} & \textbf{0.71} & \textbf{0.73} & \textbf{0.70} & \textbf{0.69} & \textbf{0.65}\\ 
\textit{w/o PE }& 0.65 & 0.45 & 0.64 & 0.46 & 0.65 & 0.45 & 0.63 & 0.50 \\ 
\textit{w/o G-MSA}& 0.59 & 0.25 & 0.55 & 0.20 & 0.51 & 0.20 & 0.45 & 0.17  \\  
\textit{w/o C-MSA }& 0.50 & 0.20 & 0.48 & 0.10 & 0.52 & 0.15 & 0.40 & 0.20\\ 
\textit{vanilla class}& 0.60 & 0.51 & 0.65 & 0.55 & 0.66 & 0.50 & 0.65 & 0.60 \\ 
\hline 
\end{tabular}
    }
    \vspace{-2mm}
    \caption{Ablation study results on DBG\_F1 (D1-D4).}
    \label{tab:ablation}
    \vspace{-2mm}
\end{table}

\subsubsection{What has C-MSA learned?}
\begin{figure}[t]
    \centering
    \includegraphics[width=\linewidth]{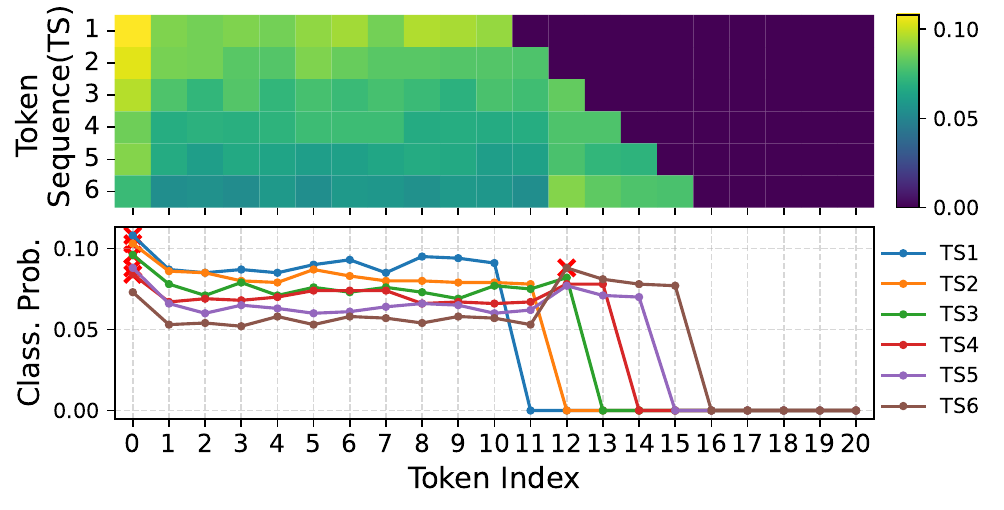}
    \vspace{-7mm}
    \caption{Attention distribution learned by C-MSA. Higher weights are assigned to tokens near drift time point and the first context token, with red `$\times$' indicating TRACE's predicted drift and the ground truth at token 12.}
    \label{fig:attn-scores}
    \vspace{-5mm}
\end{figure}

To investigate TRACE's internal mechanism, we analyze the attention patterns learned by C-MSA. 
Concretely, we visualize its attention weights on token sequences from six consecutive steps in DBG\_F1D1 (Figure~\ref{fig:attn-scores}), along with the corresponding classification scores. 
The results reveals that attention is unevenly distributed, with C-MSA consistently focuses on tokens corresponding to time steps with notable distributional shifts. 
The first token, representing the current environment, consistently receives high attention, underscoring its role as an anchor of the current environment. 
This selective focus indicates that C-MSA effectively identifies drift relevant features, contributing to both the interpretability and generalization ability of TRACE.

\subsubsection{What has G-MSA learned?}
\begin{figure}[t]
    \centering
    \subfloat[TS1]{
    \includegraphics[width=0.3\linewidth]{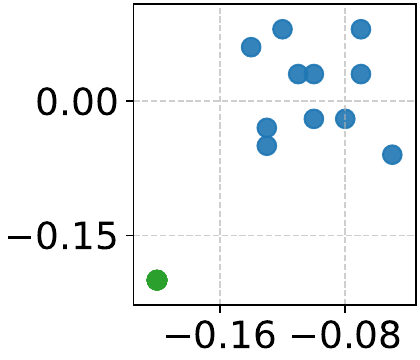}
    }
    \subfloat[TS2]{
    \includegraphics[width=0.3\linewidth]{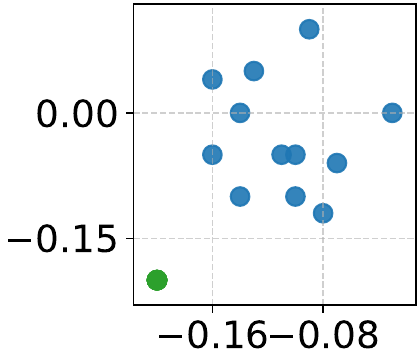}
    }
    \subfloat[TS3]{
    \includegraphics[width=0.3\linewidth]{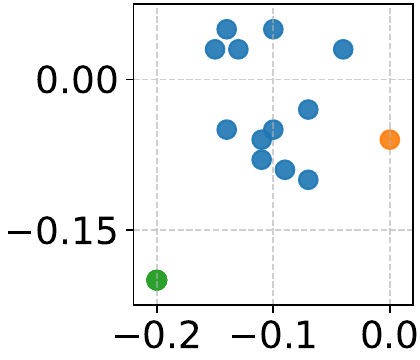}
    }\\
    \vspace{-2mm}
    \subfloat[TS4]{
    \includegraphics[width=0.3\linewidth]{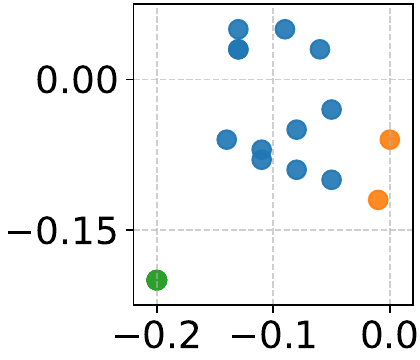}
    }
    \subfloat[TS5]{
    \includegraphics[width=0.3\linewidth]{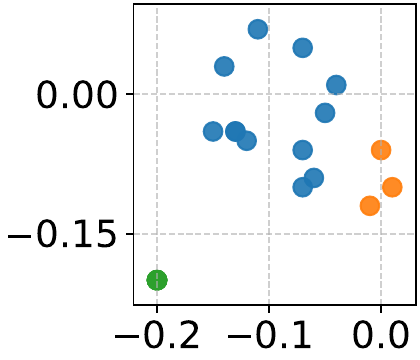}
    }
    \subfloat[TS6]{
    \includegraphics[width=0.3\linewidth]{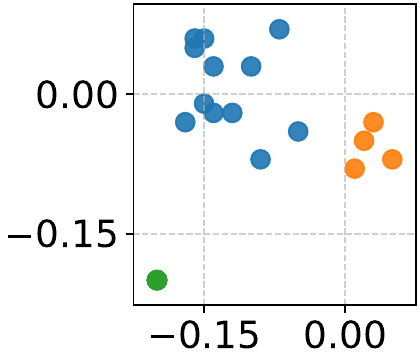}
    }
    \vspace{-2mm}
    \caption{G-MSA token distribution via PCA. Blue Points: context-similar tokens; Orange Points: distributional changes; Green Points: padding tokens.}
    \label{fig:vmsa-token-distribution}
    \vspace{-2mm}
\end{figure}

To understand the representations learned by G-MSA, we extract its output embeddings from six consecutive time steps (TS1–TS6) in DBG\_F1D1, consistent with the previous subsection. 
These high-dimensional vectors are projected into 2D using PCA~\cite{shlens2014tutorial} for visualization.
Figure~\ref{fig:vmsa-token-distribution} shows the token distributions per time point. 
Blue points indicate in-distribution tokens (aligned with the first context token), orange points denote drifted tokens, and green points correspond to padding. 
As the stream arriving, token clusters shift, and drifted tokens become increasingly separated from the main cluster.
This behavior demonstrates that G-MSA captures global sequence dependencies and distinguishes between coherent and distributionally distinct tokens. 
Such structured separation highlights G-MSA's critical role in enabling TRACE to identify drift across time.

\subsection{Application to Stream Clustering Tasks (RQ4)}
To evaluate the practical utility and generalization capability of TRACE-EA in real-world scenarios, we apply it to streaming clustering tasks. 
These tasks are fundamental in streaming environments, where maintaining clustering quality under continuous distributional changes is critical for decision-making in domains such as network monitoring \cite{borghesi2019anomaly,bulut2005unified}, energy systems \cite{yu2019stream}, and user behavior analysis~\cite{wang2016unsupervised,daza2023edbb}.
For proof-of-principled evaluation, we use
four widely-used and well-established datasets: \textit{Convtype}~\cite{blackard1999comparative}, \textit{Electricity}~\cite{asuncion2007uci}, \textit{Kddcup99}~\cite{kddcup2007}, and \textit{Pokerhand}~\cite{asuncion2007uci}.
These datasets feature diverse and well-documented drift patterns and are commonly used for evaluating stream clustering algorithms. 
More details of datasets are provided in the Appendix E.

\begin{figure}[t]
    \centering
    \captionsetup[subfloat]{captionskip=-1.5pt}
    \vspace{-2mm}
    \subfloat[Convtype]{
    \includegraphics[width=0.45\linewidth]{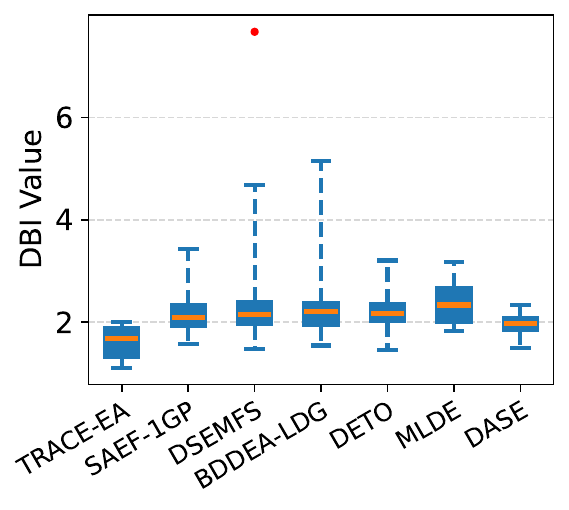}
    }
    \subfloat[Electricity]{
    \includegraphics[width=0.45\linewidth]{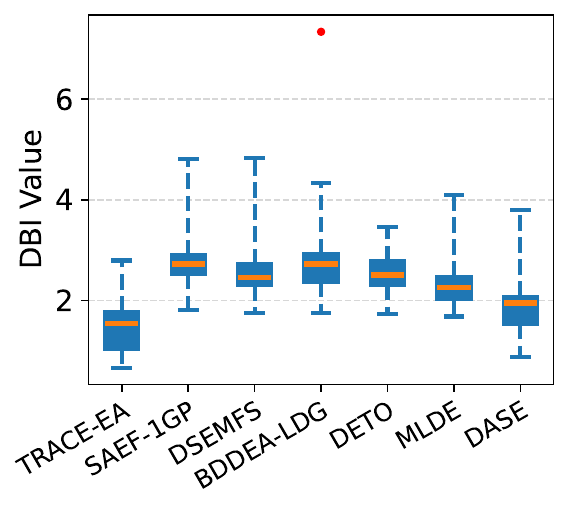}
    }\vspace{-3mm}
    \subfloat[Kddcup99]{
    \includegraphics[width=0.45\linewidth]{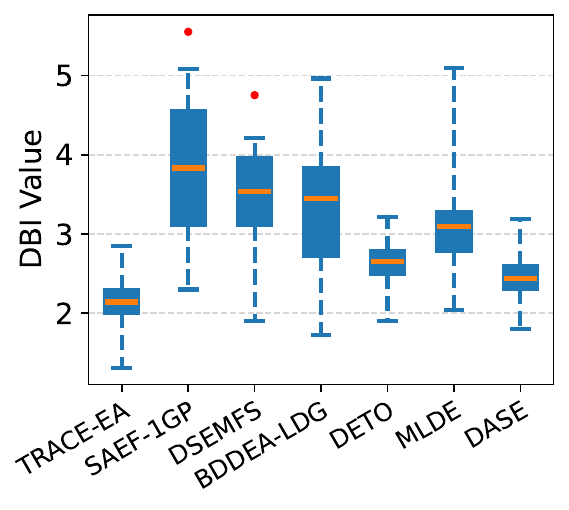}
    }
    \subfloat[Pokerhand]{
    \includegraphics[width=0.45\linewidth]{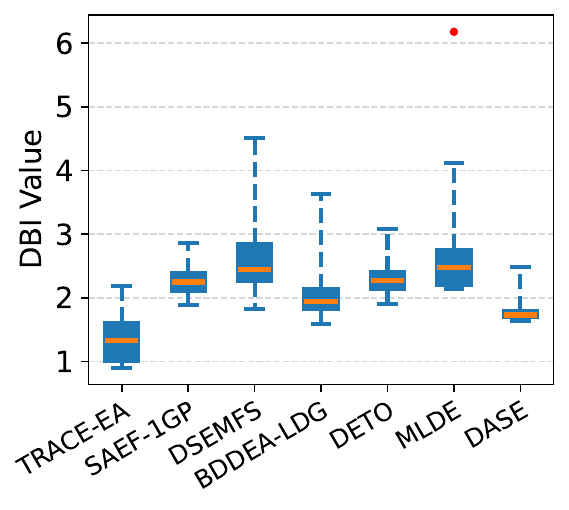}
    }
    \vspace{-2mm}
    \caption{Boxplot of DBI values obtained by TRACE-EA and SDDEAs on real-world data stream clustering datasets.}
    \label{fig:real-world-boxplot}
    \vspace{-2mm}
\end{figure}

All algorithms in the experiment are integrated into the ACDE~\cite{das2008automatic} framework to encode individuals for optimization, using the Davies-Bouldin Index (DBI) \cite{davies2009cluster} as the optimization objective. 
Lower DBI values indicate better clustering performance.
At each time point, when a new batch of data arrives, TRACE evaluates whether concept drift has occurred. 
If no drift is detected, optimization proceeds using the current state.
If drift is detected, surrogate is updated to adapt to the new environment before resuming optimization.
Figure~\ref{fig:real-world-boxplot} summarizes DBI values over 11 runs. 
TRACE-EA consistently achieves lower scores with reduced variance, demonstrating superior clustering quality and robustness. 
Improvements are especially pronounced on \textit{Electricity} and \textit{Kddcup99}, which involve varying drifts, highlighting TRACE-EA's effectiveness in dynamic real-world environments.

\section{Conclusion}
This paper presents TRACE, a novel and transferable concept drift estimator designed for detecting distributional changes in streaming data for SDDO problems.
TRACE leverages a principled stream tokenization method to extract statistical representations from data streams, facilitating learning-based drift modeling. 
The resulting labeled token sequences are processed by a dual-attention encoder to jointly captures the global context and localized drift patterns, enabling precise identification of distributional shifts. 
A pointer-like classification head then accurately pinpoints the most probable drift index within the sequence.
Building upon TRACE, we further introduce TRACE-EA, an instantiation of SDDEA that integrates TRACE as a plug-and-play drift detector. 
TRACE-EA supports generalized, drift-aware optimization across previously unseen datasets and tasks. 
Comprehensive experiments on synthetic and real-world datasets confirm that TRACE achieves accurate and robust drift detection, while TRACE-EA consistently enhances the optimization performance of SDDO. 
However, this study also has limitations. 
First, the fixed sliding window causes detection delay, which could be mitigated by adaptive windowing techniques. 
Second, a tighter integration of the detector and optimizer, beyond the current plug-and-play design, could yield further performance gains. The field of automated algorithm design offers a promising avenue for discovering such synergistic frameworks automatically~\cite{chen2024symbol,ma2025toward,guo2025designx}.

\section{Acknowledgments}
This work was supported in part by Guangdong Natural Science Funds for Distinguished Young Scholars (Grant No. 2022B1515020049), in part by National Natural Science Foundation of China (Grant No. 62276100), in part by Guangzhou Science and Technology Elite Talent Leading Program for Basic and Applied Basic Research (Grant No. SL2024A04J01361), and in part by the Fundamental Research Funds for the Central Universities (Grant Nos. 2025ZYGXZR027 and ZYTS25297).
\bibliography{ref}

@inproceedings{zhong2024sddobench,
  title = {SDDObench: A benchmark for streaming data-driven optimization with concept drift},
  shorttitle = {SDDObench},
  booktitle = {Proceedings of the Genetic and Evolutionary Computation Conference},
  author = {Zhong, Yuan-Ting and Wang, Xin-Can and Sun, Yu-Hong and Gong, Yue-Jiao},
  year = {2024},
  pages = {59--67}
}

@article{vaswani2017attention,
  title={Attention is all you need},
  author={Vaswani, Ashish and Shazeer, Noam and Parmar, Niki and Uszkoreit, Jakob and Jones, Llion and Gomez, Aidan N and Kaiser, {\L}ukasz and Polosukhin, Illia},
  journal={Advances in Neural Information Processing Systems},
  volume={30},
  year={2017}
}

@article{das2008automatic,
  title={Automatic clustering using an improved differential evolution algorithm},
  author={Das, Swagatam and Abraham, Ajith and Konar, Amit},
  journal={IEEE Transactions on Systems, Man, and Cybernetics-Part A: Systems and Humans},
  volume={38},
  number={1},
  pages={218--237},
  year={2008},
  publisher={IEEE}
}

@inproceedings{gama2004learning,
  title={Learning with drift detection},
  author={Gama, Joao and Medas, Pedro and Castillo, Gladys and Rodrigues, Pedro},
  booktitle={Proceeding  of Advances in Artificial Intelligence--SBIA 2004: 17th Brazilian Symposium on Artificial Intelligence},
  pages={286--295},
  year={2004},
}

@article{li2023datadriven,
  title = {A data-driven evolutionary transfer optimization for expensive problems in dynamic environments},
  author = {Li, Ke and Chen, Renzhi and Yao, Xin},
    year={2024},
  volume={28},
  number={5},
  pages={1396-1411},
  journal = {IEEE Transactions on Evolutionary Computation},
  issn = {1941-0026}
}

@article{yang2023data,
  title = {A data stream ensemble assisted multifactorial evolutionary algorithm for offline data-driven dynamic optimization},
  author = {Yang, Cuie and Ding, Jinliang and Jin, Yaochu and Chai, Tianyou},
  year = {2023},
  journal = {Evolutionary Computation},
  pages = {1--25}
}

@ARTICLE{luo2018surrogate,
  author={Luo, Wenjian and Yi, Ruikang and Yang, Bin and Xu, Peilan},
  journal={IEEE Transactions on Emerging Topics in Computational Intelligence}, 
  title={Surrogate-assisted evolutionary framework for data-driven dynamic optimization}, 
  year={2019},
  volume={3},
  number={2},
  pages={137-150},
  doi={10.1109/TETCI.2018.2872029}}

@techreport{li2008benchmark,
  title={Benchmark generator for CEC 2009 competition on dynamic optimization},
  author={Li, Changhe and Yang, Shengxiang and Nguyen, Trung-Thanh and Yu, E Ling and Yao, Xin and Jin, Yaochu and Beyer, HG and Suganthan, Ponnuthurai Nagaratnam},
  year={2008}
}

@article{yazdani2022benchmarking,
  title = {benchmarking continuous dynamic optimization: Survey and generalized test suite},
  shorttitle = {Benchmarking Continuous Dynamic Optimization},
  author = {Yazdani, Danial and Omidvar, Mohammad Nabi and Cheng, Ran and Branke, J{\"u}rgen and Nguyen, Trung Thanh and Yao, Xin},
  year = {2022},
  month = may,
  journal = {IEEE Transactions on Cybernetics},
  volume = {52},
  number = {5},
  pages = {3380--3393},
  issn = {2168-2275},
  doi = {10.1109/TCYB.2020.3011828}
}

@inproceedings{bifet2007learning,
  title = {Learning from time-changing data with adaptive windowing},
  booktitle = {Proceedings of the 2007 SIAM International Conference on Data Mining},
  author = {Bifet, Albert and Gavald{\`a}, Ricard},
  year = {2007},
  month = apr,
  pages = {443--448},
  doi = {10.1137/1.9781611972771.42},
  langid = {english}
}

@article{frias-blanco2015online,
  title = {Online and Non-Parametric Drift Detection Methods Based on Hoeffding's Bounds},
  author = {{Frias-Blanco}, Isvani and {Campo-Avila}, Jose Del and {Ramos-Jimenez}, Gonzalo and {Morales-Bueno}, Rafael and {Ortiz-Diaz}, Agustin and {Caballero-Mota}, Yaile},
  year = {2015},
  month = mar,
  journal = {IEEE Transactions on Knowledge and Data Engineering},
  volume = {27},
  number = {3},
  pages = {810--823},
  doi = {10.1109/TKDE.2014.2345382},
  langid = {english}
}

@incollection{pesaranghader2016fast,
  title = {Fast hoeffding drift detection method for evolving data streams},
  booktitle = {Machine Learning and Knowledge Discovery in Databases},
  author = {Pesaranghader, Ali and Viktor, Herna L.},
  year = {2016},
  volume = {9852},
  pages = {96--111}
}

@article{raab2020reactive,
  title = {Reactive soft prototype computing for concept drift streams},
  author = {Raab, Christoph and Heusinger, Moritz and Schleif, Frank-Michael},
  year = {2020},
  month = nov,
  journal = {Neurocomputing},
  volume = {416},
  pages = {340--351},
  issn = {09252312},
  doi = {10.1016/j.neucom.2019.11.111},
  urldate = {2024-10-04},
  langid = {english}
}

@inproceedings{alsaedi2023radar,
  title = {RADAR: Reactive concept drift management with robust variational inference for evolving iot data streams},
  shorttitle = {RADAR},
  booktitle = {Proceedings of IEEE 39th International Conference on Data Engineering (ICDE)},
  author = {Alsaedi, Abdullah and Sohrabi, Nasrin and Mahmud, Redowan and Tari, Zahir},
  year = {2023},
  pages = {1995--2007},
  langid = {english}
}

@inproceedings{wan2024online,
  title = {Online drift detection with maximum concept discrepancy},
  booktitle = {Proceedings of the 30th ACM SIGKDD Conference on Knowledge Discovery and Data Mining},
  author = {Wan, Ke and Liang, Yi and Yoon, Susik},
  year = {2024},
  pages = {2924--2935}
}

@article{zhang2024solving,
  title={solving expensive optimization problems in dynamic environments with meta-learning},
  author={Zhang, Huan and Ding, Jinliang and Feng, Liang and Tan, Kay Chen and Li, Ke},
  journal={IEEE Transactions On Cybernetics},
  year={2024}
}

@inproceedings{wu2024surrogate,
  title={A surrogate-assisted evolutionary algorithm for expensive dynamic multimodal optimzation},
  author={Wu, Xunfeng and Liu, Songbai and Ji, Junkai and Ma, Lijia and Leung, Victor CM},
  booktitle={Proceedings of IEEE Congress on Evolutionary Computation (CEC)},
  pages={1--8},
  year={2024},
}

@article{shlens2014tutorial,
  title={A tutorial on principal component analysis},
  author={Shlens, Jonathon},
  journal={arXiv preprint arXiv:1404.1100},
  year={2014}
}

@article{li2020boosting,
  title={Boosting data-driven evolutionary algorithm with localized data generation},
  author={Li, Jian-Yu and Zhan, Zhi-Hui and Wang, Chuan and Jin, Hu and Zhang, Jun},
  journal={IEEE Transactions on Evolutionary Computation},
  volume={24},
  number={5},
  pages={923--937},
  year={2020},
}

@article{huang2021offline,
  title = {Offline data-driven evolutionary optimization based on tri-training},
  author = {Huang, Pengfei and Wang, Handing and Jin, Yaochu},
  year = {2021},
  month = feb,
  journal = {Swarm and Evolutionary Computation},
  volume = {60},
  pages = {100800}
}

@misc{asuncion2007uci,
  title={UCI machine learning repository},
  author={Asuncion, Arthur and Newman, David and others},
  year={2007},
  publisher={Irvine, CA, USA}
}

@article{blackard1999comparative,
  title={Comparative accuracies of artificial neural networks and discriminant analysis in predicting forest cover types from cartographic variables},
  author={Blackard, Jock A and Dean, Denis J},
  journal={Computers and Electronics in Agriculture},
  volume={24},
  number={3},
  pages={131--151},
  year={1999},
  publisher={Elsevier}
}

@misc{kddcup2007,
  title={https://kdd.ics.uci.edu/databases/kddcup\\99/kddcup99.html},
  author={KddCup99},
  year={2007},
}

@inproceedings{ji2022stden,
  title={STDEN: Towards physics-guided neural networks for traffic flow prediction},
  author={Ji, Jiahao and Wang, Jingyuan and Jiang, Zhe and Jiang, Jiawei and Zhang, Hu},
  booktitle={Proceedings of the AAAI Conference on Artificial Intelligence},
  volume={36},
  number={4},
  pages={4048--4056},
  year={2022}
}

@article{zhong2025data,
  title={Data-driven evolutionary computation under continuously streaming environments: A drift-aware approach},
  author={Zhong, Yuan-Ting and Gong, Yue-Jiao},
  journal={IEEE Transactions on Evolutionary Computation},
  year={2025},
  publisher={IEEE}
}

@inproceedings{kang2019dynamic,
  title={Dynamic vehicle traffic control using deep reinforcement learning in automated material handling system},
  author={Kang, Younkook and Lyu, Sungwon and Kim, Jeeyung and Park, Bongjoon and Cho, Sungzoon},
  booktitle={Proceedings of the AAAI Conference on Artificial Intelligence},
  volume={33},
  number={01},
  pages={9949--9950},
  year={2019}
}

@inproceedings{gower2025identifying,
  title={Identifying predictions that influence the future: Detecting performative concept drift in data streams},
  author={Gower-Winter, Brandon and Krempl, Georg and Dragomiretskiy, Sergey and Jelsma, Tineke and Siebes, Arno},
  booktitle={Proceedings of the AAAI Conference on Artificial Intelligence},
  volume={39},
  number={11},
  pages={11726--11734},
  year={2025}
}

@inproceedings{lu2025early,
  title={Early concept drift detection via prediction uncertainty},
  author={Lu, Pengqian and Lu, Jie and Liu, Anjin and Zhang, Guangquan},
  booktitle={Proceedings of the AAAI Conference on Artificial Intelligence},
  volume={39},
  number={18},
  pages={19124--19132},
  year={2025}
}

@inproceedings{styler2015real,
  title={Real-time predictive optimization for energy management in a hybrid electric vehicle},
  author={Styler, Alexander and Nourbakhsh, Illah},
  booktitle={Proceedings of the AAAI Conference on Artificial Intelligence},
  volume={29},
  number={1},
  year={2015}
}

@article{kruskal1952use,
  title={Use of ranks in one-criterion variance analysis},
  author={Kruskal, William H and Wallis, W Allen},
  journal={Journal of the American statistical Association},
  volume={47},
  number={260},
  pages={583--621},
  year={1952}
}

@article{dunnett1955multiple,
  title={A multiple comparison procedure for comparing several treatments with a control},
  author={Dunnett, Charles W},
  journal={Journal of the American Statistical Association},
  volume={50},
  number={272},
  pages={1096--1121},
  year={1955},
  publisher={Taylor \& Francis}
}

@inproceedings{richter2020model,
  title={Model-based optimization with concept drifts},
  author={Richter, Jakob and Shi, Junjie and Chen, Jian-Jia and Rahnenf{\"u}hrer, J{\"o}rg and Lang, Michel},
  booktitle={Proceedings of  Genetic and Evolutionary Computation Conference},
  pages={877--885},
  year={2020}
}

@article{liu2025data,
  title={Data stream driven dynamic multiobjective optimization using surrogate transfer},
  author={Liu, Zhening and Wang, Handing and Gong, Maoguo and Jin, Yaochu},
  journal={IEEE Transactions on Emerging Topics in Computational Intelligence},
  year={2025},
  publisher={IEEE}
}

@article{davies2009cluster,
  title={A cluster separation measure},
  author={Davies, David L and Bouldin, Donald W},
  journal={IEEE Transactions on Pattern Analysis And Machine Intelligence},
  number={2},
  pages={224--227},
  year={2009},
  publisher={Ieee}
}

@inproceedings{bulut2005unified,
  title={A unified framework for monitoring data streams in real time},
  author={Bulut, Ahmet and Singh, Ambuj K},
  booktitle={Proceedings of 21st International Conference on Data Engineering (ICDE)},
  pages={44--55},
  year={2005},
  organization={IEEE}
}

@inproceedings{borghesi2019anomaly,
  title={Anomaly detection using autoencoders in high performance computing systems},
  author={Borghesi, Andrea and Bartolini, Andrea and Lombardi, Michele and Milano, Michela and Benini, Luca},
  booktitle={Proceedings of the AAAI Conference on artificial intelligence},
  volume={33},
  number={01},
  pages={9428--9433},
  year={2019}
}

@article{yu2019stream,
  title={A stream processing framework based on linked data for information collaborating of regional energy networks},
  author={Yu, Han and Da Xu, Li and Cai, Hongming and Li, Shancang and Xu, Boyi and Jiang, Lihong},
  journal={IEEE Transactions on Industrial Informatics},
  volume={17},
  number={1},
  pages={179--188},
  year={2019},
  publisher={IEEE}
}

@inproceedings{wang2016unsupervised,
  title={Unsupervised clickstream clustering for user behavior analysis},
  author={Wang, Gang and Zhang, Xinyi and Tang, Shiliang and Zheng, Haitao and Zhao, Ben Y},
  booktitle={Proceedings of the 2016 CHI conference on human factors in computing systems},
  pages={225--236},
  year={2016}
}

@inproceedings{daza2023edbb,
  title={edBB-Demo: Biometrics and behavior analysis for online educational platforms},
  author={Daza, Roberto and Morales, Aythami and Tolosana, Ruben and Gomez, Luis F and Fierrez, Julian and Ortega-Garcia, Javier},
  booktitle={Proceedings of the AAAI Conference on Artificial Intelligence},
  volume={37},
  number={13},
  pages={16422--16424},
  year={2023}
}

@article{ma2025toward,
  title={Toward automated algorithm design: A survey and practical guide to meta-black-box-optimization},
  author={Ma, Zeyuan and Guo, Hongshu and Gong, Yue-Jiao and Zhang, Jun and Tan, Kay Chen},
  journal={IEEE Transactions on Evolutionary Computation},
  year={2025},
  publisher={IEEE}
}

@inproceedings{chen2024symbol,
  title={SYMBOL: Generating flexible black-box optimizers through symbolic equation learning},
  author={Chen, Jiacheng and Ma, Zeyuan and Guo, Hongshu and Ma, Yining and Zhang, Jie and Gong, Yue-Jiao},
  booktitle={The Twelfth International Conference on Learning Representations},
  year={2024}
}

@inproceedings{guo2025designx,
  title={DesignX: Human-Competitive Algorithm Designer for Black-Box Optimization},
  author={Guo, Hongshu and Ma, Zeyuan and Ma, Yining and Zhang, Xinglin and Chen, Wei-Neng and Gong, Yue-Jiao},
  booktitle={The Thirty-Ninth Annual Conference on Neural Information Processing Systems},
  year={2025}
}

\end{document}